\documentclass{article}
\usepackage{arxiv}

\usepackage[utf8]{inputenc}
\usepackage[T1]{fontenc}
\usepackage[american]{babel}
\usepackage{csquotes}
\usepackage{graphicx}
\usepackage{float}
\usepackage{tabularx}
\usepackage{enumitem}
\usepackage[backend=biber,style=apa,natbib=true,maxbibnames=99]{biblatex}
\usepackage{microtype}
\usepackage{url}
\usepackage{hyperref}

\graphicspath{{figures/}}
\title{Nudging Decision Models: Do Digital Nudges Steer Fast, Generation-Free AI Agents?}
\date{}
\author{Sascha Kaltenpoth\\ Paderborn University\\ Paderborn, Germany\\ \texttt{sascha.kaltenpoth@uni-paderborn.de}
\And Haya Halimeh\\ Paderborn University\\ Paderborn, Germany\\ \texttt{haya.halimeh@uni-paderborn.de}
\AND Kevin Bösch\\ Paderborn University\\ Paderborn, Germany\\ \texttt{kboesch@mail.uni-paderborn.de}
\And Oliver Müller\\ Paderborn University\\ Paderborn, Germany\\ \texttt{oliver.mueller@uni-paderborn.de}}
\renewcommand{\shorttitle}{Nudging Decision Models}
\hypersetup{
  pdftitle={Nudging Decision Models: Do Digital Nudges Steer Fast, Generation-Free AI Agents?},
  pdfauthor={Sascha Kaltenpoth, Haya Halimeh, Kevin Bösch, Oliver Müller},
  pdfkeywords={Digital Nudging, Decision Models, Dual-Process Theory, AI Agents},
  hidelinks
}

\begin{document}

\maketitle

\begin{abstract}\noindent
LLM-based agents increasingly automate business and consumer tasks but are slow and expensive. Decision models, a new class that chooses among options without generating text, are far faster and less costly. Developers call them “System One models,” referencing dual process theory. We ask whether such models can be steered by digital nudges online and whether, as the theory predicts, automatic nudges affect them more than reflective ones. We adapted an online shopping experiment, previously run with humans and LLM agents, to three decision models (Jev, Decision-1, d1). Across 600 runs per model, they chose a preselected default in 97.5–99.9\% and a product with social influence information in 75–96\% of choices, versus 47–49\% in the control. All responded more strongly to the automatic nudge, the opposite of what has been reported for reasoning-capable LLM agents. We provide first evidence that non-generative decision models are susceptible to digital nudges in a System 1 pattern, implying agent-type-specific nudge effects.
\end{abstract}


\section{Introduction}
Agentic artificial intelligence (AI), in particular agents built on large language models (LLMs), increasingly automates business and consumer tasks \citep{kumar_agentic_2026}. Agents that operate the graphical user interface (GUI) of websites enable users to delegate whole processes, such as online shopping \citep{nguyen_gui_2025}, to those agents. However, LLM agents are slow and costly \citep{abhyankar2025osworldgold,kapoor2025ai}. Recently, decision models have been proposed as a new class of AI models that do not generate text but instead make choices among predefined options \citep{liquidai2026d1,srivastava2026decision1,noauthor_notitle_nodate}. Their developers refer to them as System 1 models \citep{noauthor_notitle_nodate}, in reference to the System 1 and System 2 taxonomy of dual process theory \citep{kahneman_thinking_2011}, because they decide directly without the autoregressive reasoning that characterizes LLMs. Decision models have been shown to be faster and less costly than LLMs \citep{noauthor_notitle_nodate,liquidai2026d1,srivastava2026decision1}, and early evaluations report competitive accuracy for decision making tasks \citep{deuser_evaluating_2026,lyu_decision-oriented_2026}. While these advantages make decision models a promising alternative for decision support, such as recommendation in e-commerce or sentiment analysis \citep{lyu_decision-oriented_2026,deuser_evaluating_2026}, System 1 thinking in terms of dual process theory is usually related to higher susceptibility to automatic Type 1 nudges \citep{caraban_23_2019,hansen_nudge_2013,hummel_how_2019,johnson_defaults_2003} that often populate digital environments \citep{weinmann_digital_2016}. As such, the question arises: How susceptible are agents based on decision models to digital nudges and does this susceptibility vary by nudge types? 

We address this question by adapting a shopping experiment that was originally developed for humans and has since been replicated with traditional LLM agents as experimental subjects \citep{halimeh_nudging_2026}. We run this experiment with three decision models. In 600 runs per decision model, a GUI agent was required to buy one of two products in each of six categories in a simulated online shop. The experiment employed three different conditions for the shop interface: (i) without a nudge, (ii) with a preselected default nudge (an automatic Type 1 nudge), and (iii) with a social influence nudge (a reflective Type 2 nudge).

The results show that decision models chose the default option in 97.5\% to 99.9\% of cases and the socially influenced option in 75\% to 96\% of cases. All three models were thus steered more strongly by the automatic default nudge than by the social one. Interestingly, this is the opposite of the pattern reported in experiments using high-reasoning LLM agents as subjects, which were affected by the reflective nudge more often than by the default nudge \citep{halimeh_nudging_2026}. 

Our study makes three contributions. First, we contribute to research on digital nudging and AI agents by showing that nudge susceptibility is not limited to models that generate text. Also non-generative decision models can be influenced by digital nudges when they operate in GUIs designed for humans. Second, we contribute to the emerging framing of AI agents’ decision-making as System 1 and System 2 thinking by showing that models that decide without reasoning respond most strongly to automatic Type 1 nudges, consistent with System 1 processing and in contrast to high-reasoning LLM agents. Third, we inform practice by showing that the efficiency gains of decision models come with a distinct nudge susceptibility profile, so agents built on decision models should be deployed with at least the same caution as LLM agents. The paper proceeds as follows. Section 2 presents the theoretical background. Section 3 derives our hypotheses and details the method. Section 4 presents the results, and Section 5 discusses them and outlines our next steps toward completing this research.

\section{Theoretical Background}
Dual process theory separates human thinking processes in a fast, automatic System 1 and a slow, deliberate System 2 \citep{evans_dual-process_2013,kahneman_thinking_2011}. Based on this distinction, System 1 operates intuitively and is more prone to judgment errors than the effortful, analytical System 2 processes \citep{kahneman_maps_2003}. Analogous to this account of human thinking, current LLMs can operate in two modes. They either generate output directly from the input sequence, as earlier instruction-tuned LLMs did, or they reason step by step about the task before responding \citep{huang_towards_2023,wei_chain--thought_2022}. Direct generation has been linked to System 1 thinking, as it exhibits judgment errors typical of that system \citep{hagendorff_human-like_2023}, whereas extended reasoning has been associated with System 2 thinking because it largely avoids these errors \citep{li_system_2025}.

Current agentic systems are predominantly built on LLMs that act within their environment \citep{yao_react_2023} through external tools such as databases or APIs \citep{schick_toolformer_2023}. When these agents use operating systems, applications, or websites one click at a time via a Graphical User Interface, they are referred to as GUI agents \citep{nguyen_gui_2025}. However, GUIs are designed for humans and may sometimes contain digital nudges. Digital nudges are design elements that steer choices without restricting the options or freedom of choice \citep{weinmann_digital_2016}. Drawing on dual process theory, \citet{hansen_nudge_2013} distinguish Type 1 nudges, which act on automatic processing (e.g., preselected defaults), from Type 2 nudges, which act on reflective processing (e.g., social influence information). Prior studies have shown that GUI agents based on LLMs are susceptible to diverse nudging strategies and deceptive designs \citep{allouah_what_2025,bosch_biased_2026,cherep_framework_2026}. \citet{halimeh_nudging_2026}, for example, tested whether GUI agents based on six different LLMs can be influenced by Type 1 and Type 2 nudges, comparing their susceptibility without and with reasoning as analogies to System 1 and System 2. Their study suggests that reasoning lowers the susceptibility to Type 1 nudges but raises it to Type 2 nudges, thereby shifting rather than reducing their overall susceptibility from one type to another. 

While LLM agents have been shown to accurately complete tasks delegated to them, their reliance on autoregressive text generation introduces significant latency and computational expense \citep{abhyankar2025osworldgold,kapoor2025ai}. A new class of AI models, called decision models, addresses this limitation by bypassing text generation to selecting actions directly, which in turn enables them to make decisions faster and at a much lower computational cost \citep{chen_system1-arena_2026,sharma_typedbench_2026}. Decision models thus do not generate text as traditional LLMs do, but instead interpret language, intent and context to output a choice along with a probability score for each available option \citep{chen_system1-arena_2026,sharma_typedbench_2026}. In practice, they mostly serve as lightweight decision components within larger systems, where they judge attributes of an input, score or rank items, or select the next action, for instance deciding which model or tool should handle an incoming request \citep{ling_jev_2026}. Because they decide directly without prolonged reasoning, much like the fast and intuitive System 1 thinking \citep{kahneman_thinking_2011}, they are sometimes called System 1 models \citep{noauthor_notitle_nodate}. In terms of performance, early studies report competitive results against LLMs in classification and recommendation tasks \citep{deuser_evaluating_2026,lyu_decision-oriented_2026} as well as mixed results in clinical judgment and agent decisions \citep{huang_can_2026,li_fast_2026}. Initial evidence, however, suggests that text written to steer a decision model can affect its choices \citep{noauthor_notitle_nodate-1}. Because decision models resemble System 1 thinking and could offer a faster, cheaper foundation for agents, it is important to examine whether agents based on these models are susceptible to digital nudges and whether this susceptibility varies by nudge type.

\section{Hypotheses and Method}

\subsection{Hypotheses}
Initial evidence suggests that textual interventions designed to guide decision models can affect their choices \citep{noauthor_notitle_nodate-1}. To investigate wether this extends to different types of digital nudges, we draw on \citet{halimeh_nudging_2026}, who showed that both Type 1 and Type 2 nudges steer LLM-based GUI agents, and that high reasoning weakens Type 1 but strengthens Type 2 nudges. Decision models, by contrast, do not generate any reasoning output before making a decision, yet they can process the same information from digital environments as LLMs. Since LLM agents remain susceptible to both nudge types across reasoning levels, and decision models already respond to textual interventions, we expect decision models to be susceptible to both types of nudges as well and derive the following hypothesis:

\textbf{\emph{H1:}} \emph{Decision models exhibit systematic shifts in choice behaviour when exposed to Type 1 (automatic) and Type 2 (reflective) nudges.}

Since Type 2 nudges target reflective processing \citep{caraban_23_2019,hansen_nudge_2013}, it thus influences a choice only if its information is processed and weighed against the alternatives, whereas a Type 1 nudge is effective precisely when no such deliberation occurs \citep{evans_dual-process_2013,hansen_nudge_2013}. Defaults, a prominent Type 1 nudge, exploit inertia and status quo bias \citep{johnson_defaults_2003,samuelson_status_1988} and are among the most effective nudges in human decision making \citep{hummel_how_2019}. Decision models map their input directly to a choice, without an intermediate generation step in which the content of a nudge could be deliberated \citep{noauthor_notitle_nodate,chen_system1-arena_2026,sharma_typedbench_2026}. Their processing thus resembles System 1 and is confined to the automatic route that Type 1 nudges target, whereas a Type 2 nudge, such as a social influence label, can only affect them through the learned association between its text and the option it marks. We therefore expect:

\textbf{\emph{H2:}} \emph{A decision model is more susceptible to Type 1 (automatic) nudges than to Type 2 (reflective) nudges.}

\subsection{Method}
\begin{figure}[tbp]
  \centering
  \includegraphics[width=\linewidth]{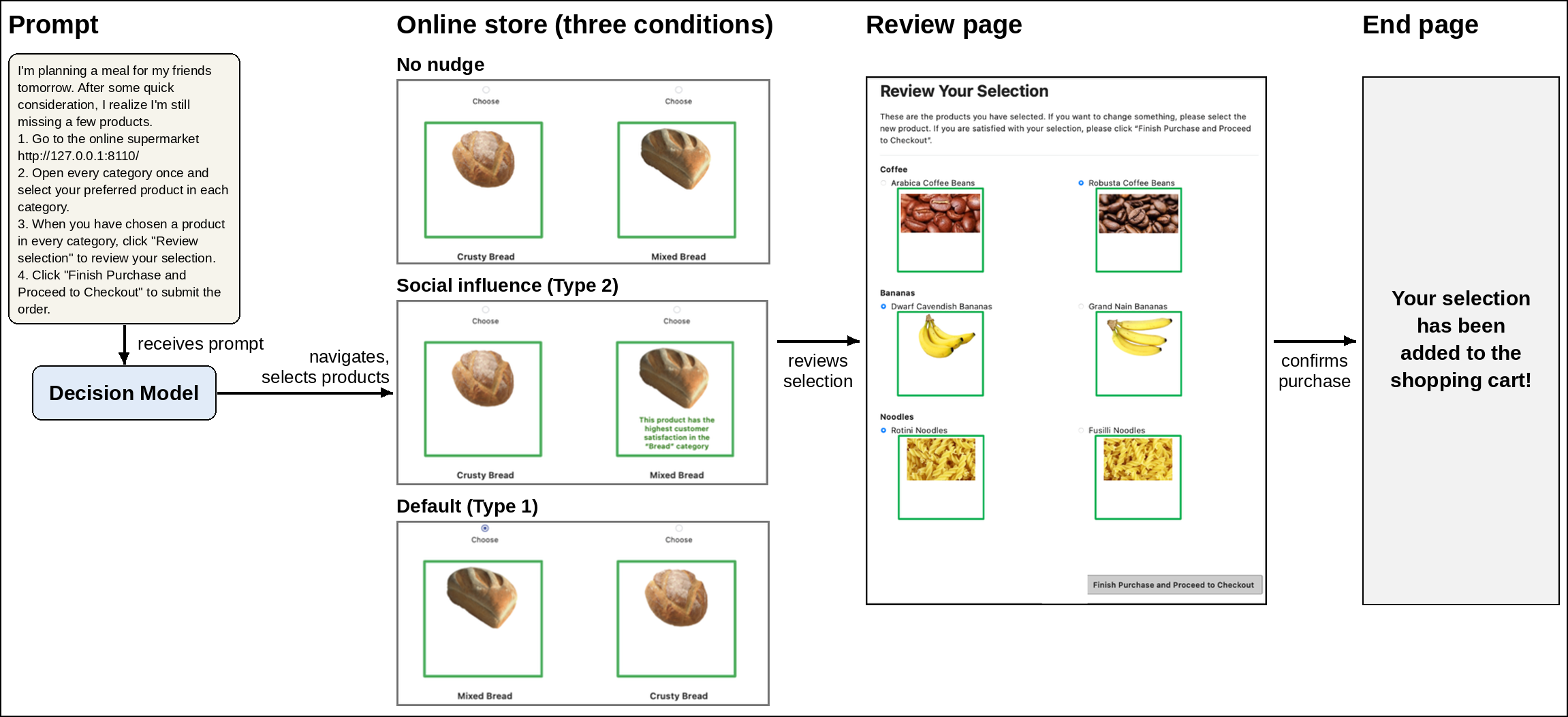}
  \caption{Procedure of a run, adapted from \citet{halimeh_nudging_2026}.}
  \label{fig:procedure}
\end{figure}

We adapt the experiment of \citet{halimeh_nudging_2026}, which replicated the human experiment of \citet{ingendahl_who_2021} with traditional LLMs, to three different decision models: Jev (\emph{jev-1.13.0}) by TypeSafe AI, Decision-1 (\emph{microsoft/microsoft-decision-1-20261009}) by Microsoft and d1 (\emph{liquid/d1-20260930}) by Liquid AI. In the study, the agent is tasked with purchasing six products in a simulated grocery store. In each product category, the agent has to decide between two products. As visible in Figure~\ref{fig:procedure}, (i) the agent has to choose one of two products in each of six categories, each shown on a separate tab, (ii) reviews its selections, and (iii) confirms the purchase. In the control condition, the store is neutral. In the default condition (automatic Type 1 nudge), one target product per category is preselected. In the social influence condition (reflective Type 2 nudge), the target carries the label ``This product has the highest customer satisfaction in the `X' category''. In half of the runs, the nudge targets one product of each pair (set A), while in the other half, it targets the other product (set B). This counterbalancing ensures that preferences for specific products cancel out. The order of categories and products was randomized in every run. We conducted 200 runs per condition (100 per set) with each decision model in random order.

To enable the decision models to navigate websites as GUI agents, we connected them to the browser through a modified version of the open-source Browser Use Library \citep{browser_use_team_celebrating_2025}. At each step, the browser interface sends the visible page text and the history of past actions to the model, along with the clickable elements on the page as options to choose from. The agent then executes the option to which the model assigns the highest probability, such as selecting a product or moving to the next tab. To ensure that the experimental design also works for decision models, we made minor adjustments to the task text and the Browser Use Implementation compared with \citet{halimeh_nudging_2026}. All changes to the original experimental design are documented in our online repository\footnote{Online repository: \url{https://anonymous.4open.science/r/nudging_decision_models-76C7/}}. Runs that exceeded 60 actions were terminated and restarted, with a maximum of three attempts per task.

Following \citet{halimeh_nudging_2026}, we estimated Bayesian multilevel logistic regressions to examine the effects of experimental conditions on the probability of choosing the target \citep{burkner_brms_2017}, with a weakly informative Student-\emph{t}(3, 0, 2.5) prior \citep{gelman2008weakly}. The model include experimental condition as a fixed effect and a random intercept per run to account for variation across runs. Results are reported as odds ratios (ORs), with values above 1 indicating increased odds of choosing the target and values below 1 indicating decreased odds. An effect is credible if the 95\% credible interval (CI) of its odds ratio excludes 1.

\section{Results}
\begin{figure}[tbp]
  \centering
  \includegraphics[width=0.85\linewidth]{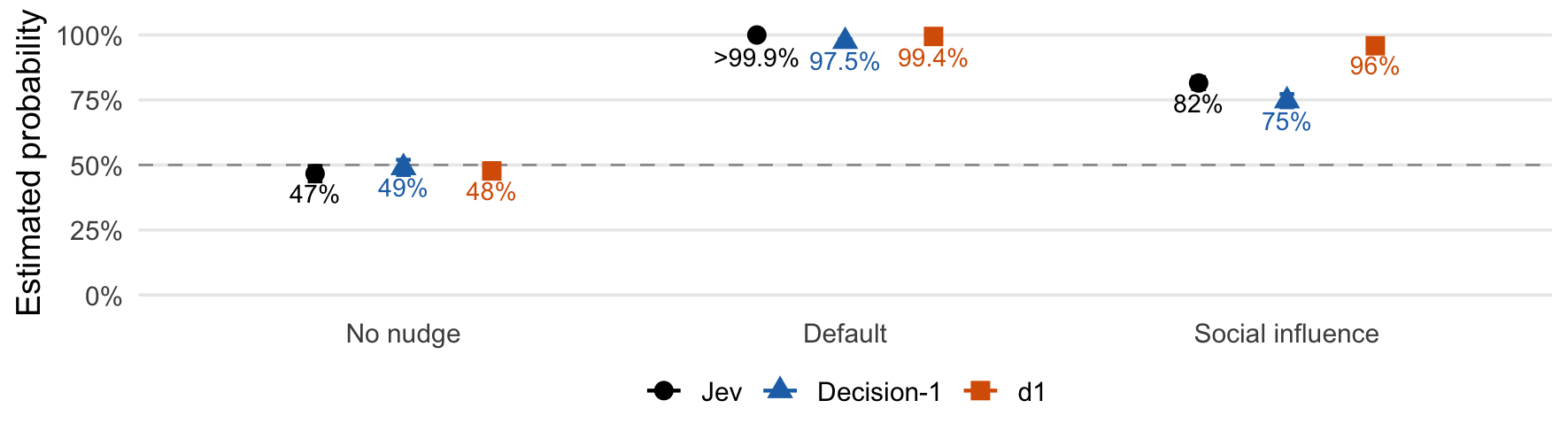}
  \caption{Bayesian multilevel model of each decision model: Estimated target product selection probabilities with 95\% credible intervals.}
  \label{fig:estimates}
\end{figure}

Figure \ref{fig:estimates} shows the estimated probabilities of choosing the target product pooled over all models (the three decision models) and runs (1,800 runs, 600 per decision model). The results indicate that both types of nudges substantially influenced the decision models' choices. In the control condition, the target product was chosen in 47\% to 49\% of cases, which suggest that absent a nudge, the models choose between the two displayed products almost at chance level. This share, however, increased to 97.5\% (Decision-1), 99.4\% (d1) and 99.9\% (Jev) under the default condition (automatic Type 1 nudge) and to 75\% (Decision-1), 82\% (Jev) and 96\% (d1) under the social influence condition (automatic Type 1 nudge).

Table~\ref{tab:models} reports the estimation results of the Bayesian multilevel logistic regression models for each decision model separately, as well as pooled over all models. The pooled analysis crosses condition with decision model and averages the condition effects across the three models. Regarding H1, the default and the social influence nudge both have credible positive effects in the pooled analysis and also for each decision model. H1 is therefore supported. Regarding H2, the difference between the default and the social influence coefficient is credibly positive in the pooled analysis (CI [2.91, 4.43]) for Jev (CI [4.67, 14.87]), for Decision-1 (CI [2.22, 2.99]) and for d1 (CI [1.27, 2.87]), supporting H2. The full estimation results for every analysis are available in the online repository.

\begin{table}[tbp]
  \small
  \renewcommand{\arraystretch}{1.15}
  \begin{tabularx}{\textwidth}{|p{3.6cm}|>{\centering\arraybackslash}X|>{\centering\arraybackslash}X|>{\centering\arraybackslash}X|>{\centering\arraybackslash}X|}
    \hline
    \textbf{Target selected} & \textbf{(1) Pooled} & \textbf{(2) Jev} & \textbf{(3) Decision-1} & \textbf{(4) d1}\\
    \hline
    Intercept & $-0.09^{*}$ (0.03) & $-0.13^{*}$ (0.06) & $-0.04$ (0.06) & $-0.09$ (0.06)\\
    \hline
    Defaults & $5.52^{*}$ (0.39) & $9.41^{*}$ (2.73) & $3.71^{*}$ (0.20) & $5.25^{*}$ (0.39)\\
    \hline
    Social influence & $1.99^{*}$ (0.07) & $1.62^{*}$ (0.10) & $1.12^{*}$ (0.09) & $3.23^{*}$ (0.16)\\
    \hline
    Std.\ dev.\ run intercept & 0.16 & 0.21 & 0.14 & 0.22\\
    \hline
    Observations (runs) & 10,800 (1,800) & 3,600 (600) & 3,600 (600) & 3,600 (600)\\
    \hline
  \end{tabularx}
  \par\footnotesize Posterior means (log-odds) with posterior standard deviations in parentheses. $^{*}$ 95\% credible interval excludes zero.
  \caption{Bayesian multilevel logistic regression results: pooled model and each decision model.}
  \label{tab:models}
\end{table}

The results furthermore show that, while all three decision models are highly responsive to the default nudge, their responses differ when it comes to the social influence nudge. Under social influence, d1 was substantially more likely to select the target product than the other two models. Its odds of choosing the target were 5.2 times those of Jev (95\% CI [3.7, 7.0]) and 7.7 times those of Decision-1 (95\% CI [5.5, 10.3]). Robustness checks for position bias (the models' preference for the first-placed product) and for product-specific effects, reported in the online repository, confirm that neither distorts the comparison between conditions in this study, as product order and target products were balanced by design.
Furthermore, across all 1,800 runs, a completed run took on average 5.8 seconds (SD 1.9) and cost USD 0.0012 (SD 0.0005) at list prices, thereby requiring substantially less computation time and lower costs than the LLM agents examined by \citet{halimeh_nudging_2026}. The effort per decision model is reported in the online repository.

\section{Discussion and Outlook}
This study examined whether decision models, which do not generate textual output but choose among possible options, are susceptible to digital nudges when they act as GUI agents on a website designed for humans. Across the three decision models, the target was selected in 97.5\% to 99.9\% of cases under the default (automatic Type 1) nudge and in 75\% to 96\% of cases under the social influence (reflective Type 2) nudge, compared with only 47\% to 49\% in the control condition without a nudge.
Decision models are on average 8 to 19 times faster and 60 to 280 times less costly than LLMs, but appear to share the same susceptibility to nudges as LLMs have.

Furthermore, all three decision models were more susceptible to Type 1 than to Type 2 nudges. This finding contributes to research on dual process theory and AI agents by showing that the System 1 analogy also applies to models specifically designed to make decisions without explicit reasoning. Whereas \citet{halimeh_nudging_2026} used the reasoning configuration (low vs. high) of LLMs as an analogy to System 1 and System 2, our results show that decision models exhibit a pattern that dual process theory predicts for automatic processing. This also sets decision models apart from LLM agents with reasoning. In the experiment of \citet{halimeh_nudging_2026}, high reasoning lowered the estimated share of agents that kept the preselected default from 92\% to 86\% but raised the share that followed the social influence label from 96\% to 98\%. Decision models, by contrast, show the reverse order. Which type of nudge is more effective therefore depends on how the agent processes the information it is provided, which is in turn shaped both by the underlying model and, where applicable, its reasoning configuration. For the delegation of tasks to agents \citep{baird_next_2021}, this means the underlying model also influences the delegated decision. An agent based on a decision model may be less influenced by information presented on a website but more likely to follow choices preselected by the interface.

Naturally, this study is scoped by its limitations.
First, we studied three closed-source decision models and one nudge of each type, and do not yet compare decision models to the LLM agents of \citet{halimeh_nudging_2026}, as implementation and task text differ slightly between the studies.
These limitations lead to our next steps. As the corpus of decision models is growing, we will extend the experiment to more of these models, especially open-source ones. Further, we will apply more Type 1 and Type 2 nudges \citep{caraban_23_2019} to test decision models' susceptibility to other nudges. This will show whether agents based on decision models generally follow automatic more than reflective nudges. Finally, comparing decision models and LLM agents under identical experimental conditions would better clarify the trade-off between computational efficiency and susceptibility to digital nudging. Ultimately, this would help establish how far interface design shapes decisions delegated to fast, generation-free agents.

\printbibliography

\end{document}